# Optimizing Design Verification using Machine Learning: Doing better than Random

William Hughes[#], Sandeep Srinivasan[#], Rohit Suvarna[#]

[#]INZONE.AI   Palo Alto, CA, USA
`{billh, sandeep , rohit} @inzone.ai`

Maithilee Kulkarni[*]

[*]Xilinx, INZONE.AI , UT-Austin
`(maithileekool@gmail.com)`

**ABSTRACT**

As integrated circuits have become progressively more complex, constrained-random stimulus [1][2][3] has become ubiquitous as a means of stimulating a design's functionality and ensuring it fully meets expectations. In theory, random stimulus allows all possible combinations to be exercised given enough time, but in practice with highly complex designs a purely random approach will have difficulty in exercising all possible combinations in a timely fashion. As a result it is often necessary to steer the Design Verification (DV) environment to generate hard-to-hit combinations. The resulting constrained-random approach is powerful but often relies on extensive human expertise to guide the DV environment in order to fully exercise the design.  As designs become more complex, the guidance aspect becomes progressively more challenging and time consuming often resulting in design schedules in which the verification time to hit all possible design coverage points is the dominant schedule limitation.

 This paper describes an approach which leverages existing constrained-random DV environment tools but which further enhances them using supervised and reinforcement machine learning. This approach provides better than random results in a highly automated fashion thereby ensuring DV objectives of full design coverage can be achieved on an accelerated timescale and with fewer resources.

  *Keywords*— **Machine Learning, Reinforcement Learning, IC Design Verification, Software Verification, Android Mobile App Verification, Network Verification, Functional Coverage, Directed Random Testing, Stimulus Generation, RISCV Processors, Microprocessors**

## I. INTRODUCTION

The first part of this paper presents an overview of the challenges in verifying complex IC's such as a Microprocessor, GPU or SOC. The second part of the paper demonstrates real-world examples of using Machine Learning to improve functional coverage thereby achieving results better than constrained-random techniques.

Two hardware verification examples are presented, one of a *Cache Controller* design and one using the open-source *RISCV-Ariane* [5] design and Google's RISCV Random Instruction Generator [6][7]. We demonstrate that a machine-learning based approach can perform significantly better on functional coverage and reaching complex hard-to-hit states than a random or constrained-random approach.

Software and hardware systems are playing an increasingly greater role in our daily life. Thus verifying these complex systems and ensuring their safe behaviour is becoming significantly more important. According to a recently published DARPA report [4], the cost of verifying an IC is approaching greater than half of the total cost of design. The cost of verification in the software segment is also increasing exponentially.

We conclude the paper with future possibilities of the application of our technology to other domains such as Software verification of mobile apps.

## II. DV Coverage and Constrained Random Verification

Design verification (DV) of integrated circuits typically involves generating stimulus of input signals and then evaluating the resulting output signals against expected values. This allows design simulations to be conducted to determine whether the design is operating correctly. Simulation failures indicate that one or more bugs are present in the design and hence the design must be modified in order to fix the bug and the simulation(s) are then re-run to verify the fix and uncover more bugs.

Writing specific tests by hand is no longer sufficient to verify all possible functionality of today's complex chip designs. In order to fully exercise all possible combinations, a constrained-random approach is typically employed whereby input stimulus combinations and sequences are generated randomly but with an extensive set of environment-controls to allow the simulation to be steered in ways which allow for a rich and diverse set of input stimulus to be generated.

However, passing some fixed set of simulations is insufficient to demonstrate that a design is free from bugs. It is also necessary to determine whether the set of simulations run on the design are sufficient to fully cover the entirety of the functionality of the design required to satisfy the design objectives. The extent to which a set of simulations covers desired design functionality is termed coverage and there are a number of different methods by which coverage can be measured and tracked.

A commonly used simple method for tracking coverage involves determining whether each line of code in a design evaluates both to a '1' and a '0' at least once each during the set of simulations. However this method can be insufficient to guarantee full coverage as there may be conditions which are deemed necessary for the design to handle for which there may not be a corresponding line of code which exactly encapsulates the condition. Common examples are cases where there may be a confluence of two or more conditions which can happen in the same cycle. Each condition may have a corresponding line of code but there may not be a line of code which fully expresses all conditions occurring simultaneously. It may also be necessary to generate certain sequences over time which again may not have corresponding design code which captures the objective.

For example, a cache memory in a microprocessor often has a limitation in the number of read/write operations it can handle simultaneously. The number of potential read/write operations will often exceed this limit. For example, the cache may be able to process two accesses in a single cycle whereas there may be read/write requests from one or more load instructions, a store buffer, a victim buffer and a snoop (cache coherence) queue. In this case there is often arbitration logic to determine which accesses proceed to the cache and which are stalled to be handled in subsequent clock cycles. To verify the design all combinations of simultaneous cache access requests must be tested including all requesters active in the same clock cycle.

In order to handle these cases it has become increasingly common to encapsulate a functional condition involving a number of existing signals in the design in a functional coverage statement. Each functional coverage statement typically represents a condition which the design must be able to handle functionally but for which no signal is already present in the design. Many such functional coverage statements may be written in order to capture cases which the design must be able to handle. The set of functional simulations run on the design must then be able to ensure that all functional coverage statements are hit at least once during the set of simulations.

## III. Design Verification Problem

As IC designs have become increasingly complex, it has become common for the list of functional coverage statements to become significantly large. Moreover with increasing complexity many of the functional coverage statements represent conditions which occur very rarely in simulation. For example if a functional coverage statement represents a set of unrelated events to all occur together in the same clock cycle and each of the events is itself rarely hit in simulation the resulting functional coverage statement will be exceedingly rare.

Since all functional coverage statements must be "hit" in order to ensure complete verification coverage of the design, each functional coverage statement which is not hit must be analyzed in order to determine why it was not hit. The simulation input parameters may then need to be tuned in order to improve the chances of hitting the coverage statement condition and the set of simulations re-run. This can be a very difficult and time consuming process which requires significant verification team resource and which can both delay the completion of design verification as well as increasing the possibility of taping out a design with an undiscovered bug since generating large numbers of difficult to hit cases improves the chances of hitting a rare design bug prior to manufacturing silicon (tapeout).

Tuning of the simulation environment in order to make a rare condition more likely to be hit includes the following:

1. Adjusting verification parameters in order to adjust input conditions (for example generating more of a particular type of transaction or series of transactions which are more likely to lead to the particular case of interest).

2. Adjusting verification responses to design outputs (for example delaying or speeding up simulation environment responses to design requests for data).
3. Modifying configuration parameters of the design (for example enabling or disabling design modes or features which may influence how likely certain internal conditions are to occur).

IV. THE SOLUTION - USING ML ALGORITHMS TO STEER THE DV ENVIRONMENT

Machine learning can greatly improve the means by which full coverage is achieved in such cases by providing a mechanism by which coverage may be tracked, verification control parameters monitored and the interaction between them learned and then improved.

The means by which this is achieved is as follows:

- A set of simulations is run on the design in which simulation and/or design parameters which may be expected to influence coverage are varied randomly. Examples of such parameters are included in the numbered list of the prior section.
- For each individual simulation run the following set of verification input data is captured:
    o Verification parameters controlling input parameters
    o Verification environment parameters
    o Design configuration parameters
    o Coverage results listing all functional coverage statements and whether each one was hit or not during the simulation
- The results from all simulations are then run through a machine learning algorithm which tracks coverage results as a function of verification input data.
- Over a series of a number of sets of simulations the ML algorithm learns which combinations of subsets of verification input data are required in order to hit each functional coverage statement. The list of functional coverage statements tracked by the ML algorithm may be filtered to exclude functional coverage statements which are hit frequently in order to focus on the set of extremely rare and difficult to hit cases
- Once trained, the ML algorithm generates a set of verification input data recommendations which will increase the chances of hitting functional coverage statements

This ML approach may be used to greatly increase the chances of hitting rare functional coverage statements thereby accelerating the process by which a design is deemed fully verified as well as increasing the probability that hard to hit bugs will be uncovered earlier in the design process.

V. NEAR MISS TRACKING FOR VERY HARD TO HIT CASES

The algorithm described in the prior section relies on the fact that all functional coverage statements will be hit at some point over a sufficiently large number of simulation runs. ML then ensures they are hit more often. This will typically be the case even for quite rare events when multiple simulation runs are run regularly (e.g. each night) over a relatively long timeframe (e.g. several weeks).

However in some cases a functional coverage statement may represent a confluence of events which themselves are so rare that the chances of all occurring in the same cycle are vanishingly small and as a result may not occur over the duration of a typical design project even with a rigorous verification environment with an extensive dedicated datacenter.

In such cases it may be useful to extend the ML algorithm described in the prior section in order to track "near misses" where a particular functional coverage statement represented by the AND of a number of signals is never – or very rarely – hit but for which the individual terms in the AND'ed list of signals are hit. This may be termed a **"near miss"**.

In addition to tracking the list of functional coverage statements, the ML algorithm identifies hard to hit functional coverage statements and expands those statements in order to additionally track any component signals which may be AND'ed together. These component signals are then tracked by the ML algorithm in the same way as described in the prior section.

After sufficient training this will therefore result in the component signals of the hard to hit functional coverage statements being hit more frequently as the ML algorithm adjusts the verification input data. The result will be a greatly increased likelihood of hitting the very rare full functional coverage statement in subsequent verification runs.

VI. EXTENSIONS TO INCREASE THE LIKELIHOOD OF HITTING BUGS

The method described thus far is likely to increase the chances of discovering bugs earlier in the design cycle since it increases the chances of hitting functional coverage statements which are often identified by design/verification teams as being tricky cases to deal with or cases deemed to be interesting, hard to hit, or unusual in some way. This has the effect of improving the richness and depth of DV simulations by helping to steer the verification stimulus into increasingly obscure corners of the design thereby uncovering bugs quickly.

However the method described thus far can also be extended to directly target bugs by treating a simulation failure due to a bug in the same way as a functional coverage statement. In this way the ML algorithm will learn combinations of verification input data which result in the discovery of a design bug and thereby, over time, steer the DV environment in ways which make discovery of further bugs more likely.

For example: if the combination of certain design modes being enabled together with a certain DV environment parameter causes a section of the design to be exercised more often, then if that section of the design happens to be buggy then we will be more likely to find a bug if the DV environment steers simulations in that direction.

Over time the buggy section of the design is gradually improved until no more bugs are found at which point the ML environment learns to steer the DV environment in a different and more fruitful direction.

VII.    MACHINE LEARNING MODELS

This section describes the Simulation environment with Machine Learning (ML) that was used to generate the input settings for the *devices under test*. The *Cache-Controller* and the *RISCV-Ariane*[5] design, were setup in a simulation environment as depicted in Figure 1.

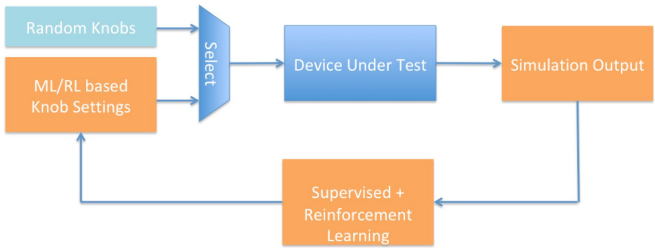

Fig. 1 Design Verification Simulation Flow using Machine Learning

The first input to the simulation environment are *Random* input configuration parameters ( *knob settings* ) that drives the simulation of the device under test. The output of the simulation based on the random knob settings is fed into a Deep Learning Neural net. This neural network , once trained , generates new knob settings that act as the next input to the simulation environment, and this process continues.

We used multiple ML architectures to generate the sequence of input knob settings for the two test cases described in this paper.

SUPERVISED LEARNING

The supervised learning based ML algorithm works by first creating a simulator Neural Network as a function approximator [10][11][17]. Given an initial random setting of input knobs and their corresponding output, the Neural Network fits a function to map the relation between the input knobs and the output. Once trained the Neural Network simulates the output for any new unseen setting of input knobs.

The objective now is to explore the input state-space of this function to find the points of maxima at the output.

In this paper we use a random search, to find the inputs that correspond to the highest values of the output (*FIFO depth or RISCV victim buffer count, described in section VIII and section IX*) as predicted by the simulator neural network. We then use these knob settings as input stimulus to the next iteration of the simulator. The simulator then produces a new set of outputs that is used to identify the ground truth values of the output and this process continues.

This process is terminated when there is a state of minimum entropy [18] and no information gain.

REINFORCEMENT LEARNING

Reinforcement Learning (RL)is an area of machine learning applicable to domains which can be modelled as an interaction between an agent that is trying to learn the optimum behavior (best action) to optimize a reward variable within a certain environment. RL has been quite successfully applied in popular games like Chess, Go [13] and DOTA [14] achieving a level of performance far beyond any human.

In an RL setup of an agent in an environment, the agent uses a trial and error approach, where it selects an action and then receives feedback in terms of 1) a reward/penalty for that action , 2) the new state of the environment as a result of the action. (See Figure 2 below)

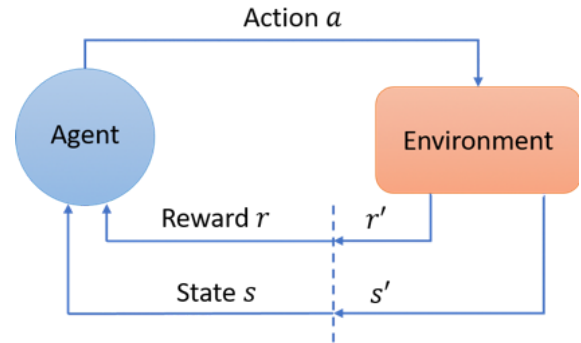

Fig. 2  RL Agent and Environment interaction

The specific RL algorithm of interest is Q-Learning. In Q-Learning the goal is to maximize the cumulative discounted reward which is given by the following Bellman Equation.

$Q\ (\ s_t,\ a_t) = E\ [\ r_{t+1} + \gamma\ r_{t+2} + \gamma^2\ r_{t+3} + ...|\ s_t,\ a_t\ ]$

for a given state *s*, and action a at time *t*, with discount factor $\gamma$.

These reward values can be stored in a lookup table or Q-Table when the problem has a limited state size and small action space. However this approach falls short for more complex problems.

Q-Learning [10] when applied to very large action and state spaces uses a DNN (Deep Neural Network) instead of a finite table lookup which is called a DQN (Deep Q-Network) [12]. The DNN acts as a Q-value function approximator and learns the Q-values for each action at a given state once trained.

In addition to the DNN an additional experience buffer which contains experiences ( *(s, a, r, s`) tuples*) are stored which are then randomly sampled to train the DQN. This is to avoid learning from consecutive samples which may have strong correlations between them.

In addition to the random search method on a simulator neural network , we can use a DQN (Deep Q-Network) to find the best set of input knobs (stimulus).

All the possible combinations of the input knobs can be considered the action space for the problem, which can be modelled as a single step Markov Decision Process, where the RL agent will learn the best policy to maximize the rewards (*FIFO depth or RISCV victim buffer count, described in section VIII and section IX*).

VIII. CACHE CONTROLLER DESIGN - EXPERIMENT #1

The first experiment was a Cache Controller design shown in Figure 3. The controller supports up to four CPU ports. CPU transactions received over the input ports go through an arbitration stage before being output to up to four separate caches. Input transaction collisions are checked at the input stage where four FIFOs - one per CPU port - hold transactions which fail collision checks with transactions from other CPU ports and hence must be serviced later.

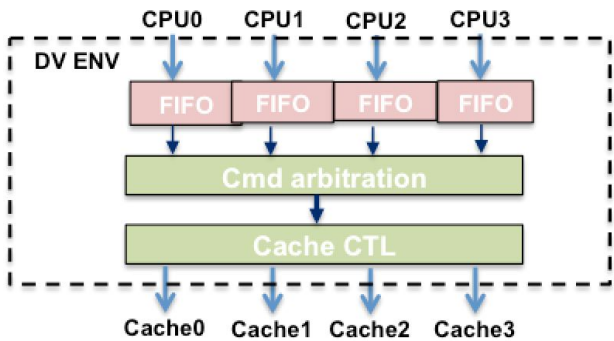

Fig. 3 Cache Controller with 4 CPU ports

This design was chosen as a proof of concept example as the input FIFOs are only allocated in the event of an incoming transaction address conflict which tend to be rare in a random transaction stream with random addresses. Also, even when the FIFO is loaded in the event of a conflict the lifetime of a transaction in the FIFO tends to be short since the FIFO entry will be de-allocated as soon as it can be processed and sent to the cache in a cycle with no address conflicts with other CPU transactions. As a result, in the presence of a stream of random input transactions with random addresses the FIFOs tend to have very low occupancy. This represents a common issue for DV where a queue/buffer can be so difficult to fill in the presence of random traffic that it seldom or perhaps never reaches a full state even across a very large number of simulations. This has implications for DV coverage since the FIFO must be filled in order to ensure correct operation. Filling the FIFO on a regular basis is also desirable in order to uncover bugs related to upstream back pressure on the transaction flow resulting from a FIFO full condition.

A suitable DV environment was developed for the cache controller in which transactions of various types could be generated on each CPU port and the corresponding output port transactions checked for correct functionality. The DV environment supported constrained-random verification as described in section II. In particular, suitable controls were provided to vary the relative weights of CPU transaction types and also to constrain transaction addresses in ranges corresponding to the tag, index and offset address of the target caches. Each simulation run involved randomly selecting values for these DV environment controls followed by a simulation involving playing a transaction stream generated based on the controls against the cache controller design and recording the resulting occupancy of the input FIFOs.

After a suitable number of simulation runs were completed, the resulting DV control settings and the corresponding FIFO occupancy simulation results were fed to a DNN (Deep Neural Network - see prior section) which, once trained, generated a set of recommended settings to use in a subsequent set of runs. After completing this second set of simulation runs using the settings from the ML agent, the results were again fed into the ML agent which generated a subsequent improved set of recommended settings. This process is summarized in Figure 4 below.

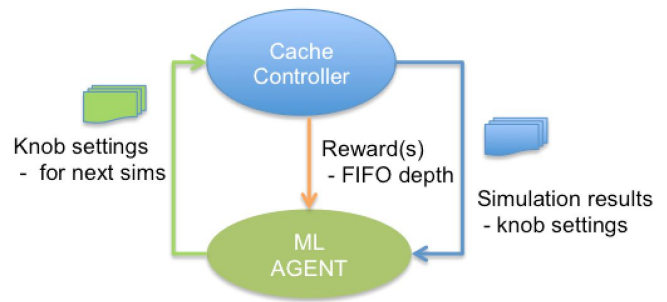

Fig. 4 Cache Controller Design ML Flow

Figure 5 below shows the resulting FIFO occupancy for each subsequent iteration (red curve). Note that the ML agent quickly learns how to adjust the DV environment control settings in order to maximize the FIFO occupancy.

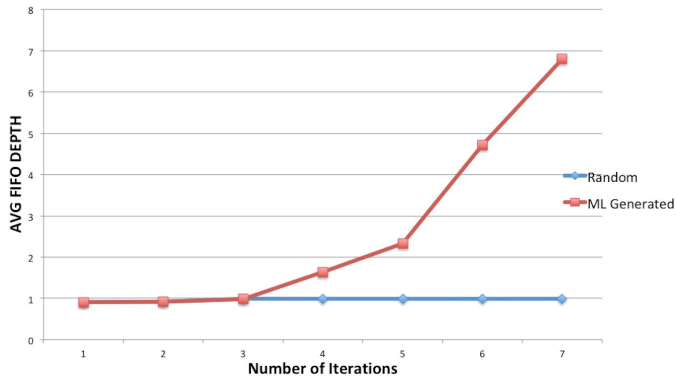

Fig. 5 Cache Controller Average FIFO Depth compared to Random

A set of runs using purely random settings without the benefit of ML-based stimulus is also shown for comparison (blue curve).

Histograms for the percentage of simulation cycles during which the FIFOs were full are shown below both for purely random (Figure 6) and with ML stimulus (Figure 7). Note the trend to higher FIFO full percentages after ML generated stimulus.

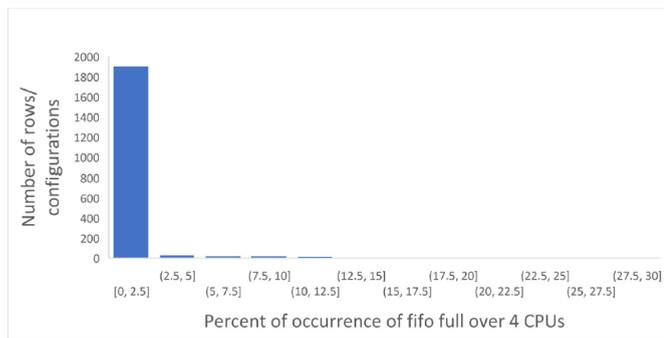

Fig. 6 Cache Controller Percent of occurrence of FIFO Full using Random

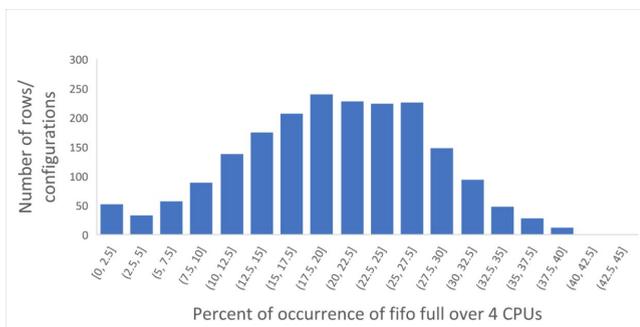

Fig. 7 Cache Controller Percent occurrence of FIFO Full using ML generated stimulus

Figure 8 below shows the percentage occurrence of FIFO full as a function of ML optimization iteration number. ML optimization is seen to significantly increase the percentage occurrence of FIFO full after just a few iterations.

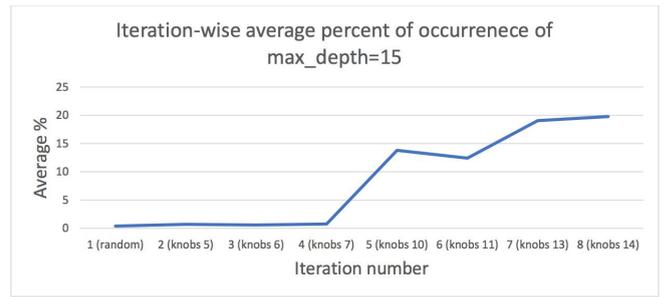

Fig.8 Cache Controller Percent occurrence of FIFO Full using ML

Figure 9 shows the average FIFO depth reached across 4 CPU's using randomly generated stimulus settings. The histogram shows that the majority of the distribution resulted in a low average FIFO depth.

Figure 10 shows the average FIFO depth across 4 CPU's using ML generated knob settings. This indicates a significant increase in the occurrence of high FIFO depths.

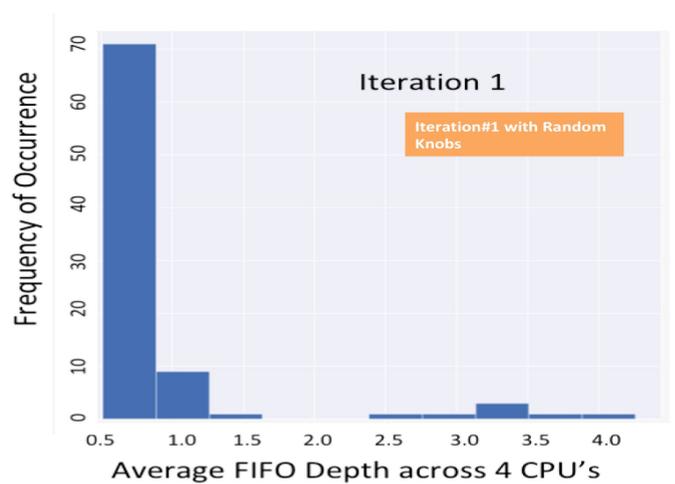

Fig. 9 Average FIFO Depth with Random Stimulus

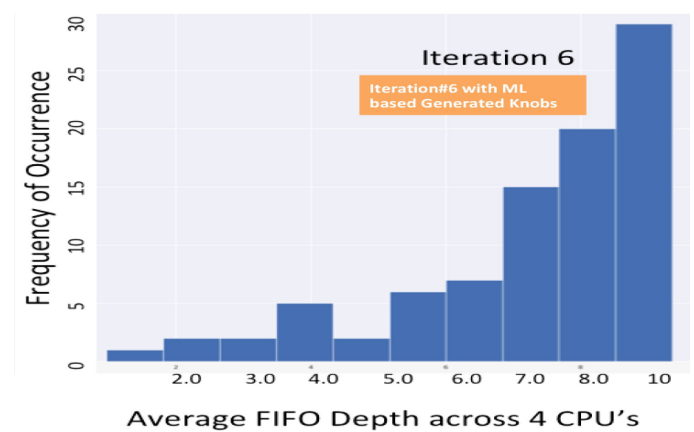

Fig.10 FIFO Depth on Iteration-6 using ML generated stimulus

The results shown in Figures 4 through 10 demonstrate how an ML-based approach can quickly and automatically optimize design coverage parameters resulting in better design coverage and accelerated stress testing of a design.

IX. RISCV-ARIANE DESIGN - EXPERIMENT #2

The second experiment was the open source RISCV-Ariane 64-bit Design [5][8] shown in Figure 11 and 12. This is a CPU core design comprising a front end fetch and decode including ICACHE, branch prediction, out-of-order execution, load-store unit including DCACHE, TLBs and exception handling.

**ARIANE: Linux Capable 64-bit core**

- **Application class processor**
- **Linux Capable**
  - M, S and U privilege modes
  - TLB
  - Tightly integrated D$ and I$
  - Hardware PTW
- **Optimized for performance**
  - Frequency: > 1.5 GHz (22 FDX)
  - Area: 185 kGE
  - Critical path: ~ 25 logic levels
- **6-stage pipeline**
  - In-order issue
  - Out-of-order write-back
  - In-order commit
- **Branch-prediction**
- **Scoreboarding**
- **Designed for extendability**

Fig. 11  RISCV-Ariane Core Specification: **Source** [8]

This design was chosen for experiment #2 in order to determine how the ML techniques described thus far - and demonstrated in experiment #1 - scale to significantly more complex designs. This RISCV design had a cache size of 2MB and a total number of lines of code (verilog) of around 78k, which could be synthesized into approximately 385 Mega Transistors (185K Gate Equivalent) [8][9].

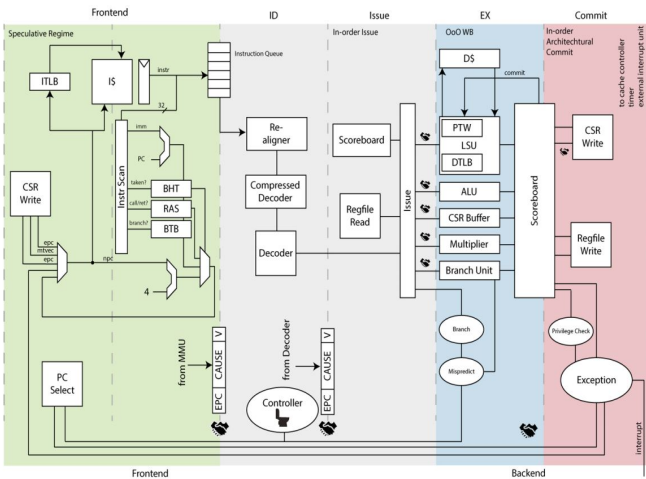

Fig.12  RISCV-Ariane block diagram : **Source** [8]

The load-store unit (LSU) has an accompanying data cache (D$) which stores recently accessed data lines from memory. The D$ has an associated *victim buffer* (VB) which temporarily stores modified lines evicted from the D$ prior to writing the modified data back to main memory. The lines are evicted when a new cache line is brought into the D$ which then replaces a modified line which is currently stored. A modified line is a cache line which has been written to within the cache but whose corresponding update(s) have not been reflected in main memory. As a result, when a modified line is evicted from the D$ it must be written back to main memory and the VB holds a number of such lines until the write has been completed.

The process of evicting a modified line from a D$ and successfully writing it back to main memory in the presence of subsequent cache accesses and external snoops (cache coherence activity generated on behalf of other system agents attempting to access the same data) can be complex and is sometimes subject to design issues/bugs.

As a result, it is highly desirable for DV to ensure a large number of victims are generated during a simulation and that the victim buffer occupancy is high in order to properly stress the design and ensure it is free of bugs in this area. This makes optimization of victim buffer occupancy in the RISCV design an ideal proof-of-concept experiment for ML based stimulus generation.

A suitable DV environment was developed for the RISCV design. An open-source instruction generator (IG) [7] was used to generate pseudo-random instruction sequences which were then used to drive simulations. A set of DV controls associated with the IG provided weights which control the relative percentage of load-store instructions versus all other instructions in combination with weights controlling the relative percentage of different load-store instruction types. These instruction control weights combined to provide a set of controls for varying instruction types while still allowing for randomization within the bounds of the relative weights specified. In addition, a set of controls were developed which constrained transaction addresses in ranges corresponding to the tag, index and offset address of the D$.

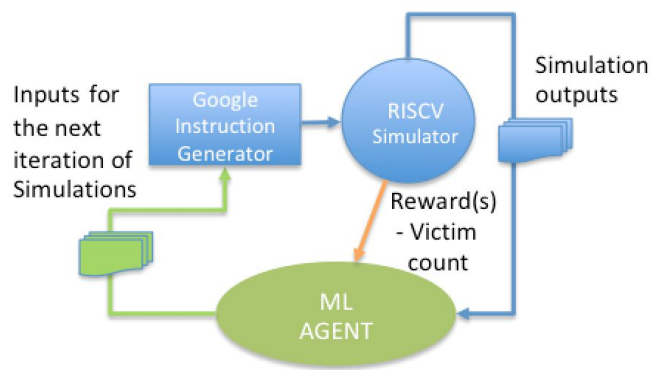

Fig. 13  RISCV and Google Instruction Generator with ML generated knobs

Using this DV environment the effectiveness of ML generated settings could be compared with the standard random approach using a similar approach to that previously described in section VIII and summarized in Figure 13.

For each individual simulation a set of IG controls was generated randomly and used by the IG to generate a constrained-random instruction sequence which was then used to drive a simulation run. The average occupancy of the D$ victim buffer was monitored during the simulation run.

After a number of simulation runs were completed, the resulting DV control settings and the corresponding VB occupancy simulation results were fed to an ML agent which, once trained, generated a set of recommended settings to use in a subsequent set of runs. This process was then repeated with each new set of simulations providing an additional iteration through the ML generated settings.

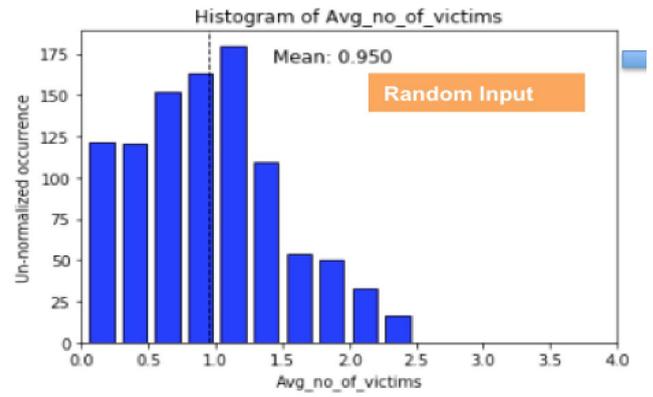

Fig.15  Google Random Instruction for the RISCV

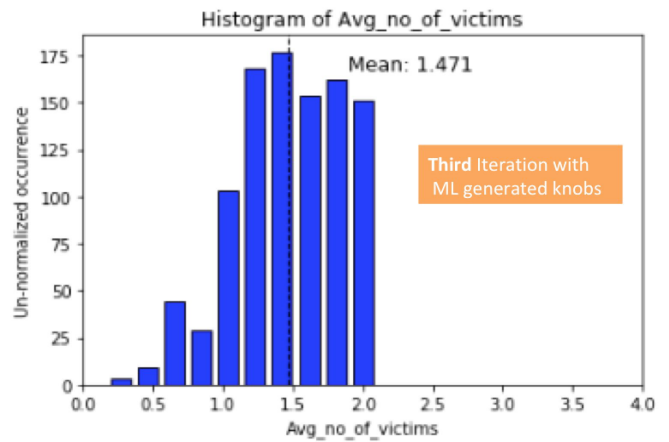

Fig.16 Google Instruction Generator with ML generated settings for RISCV

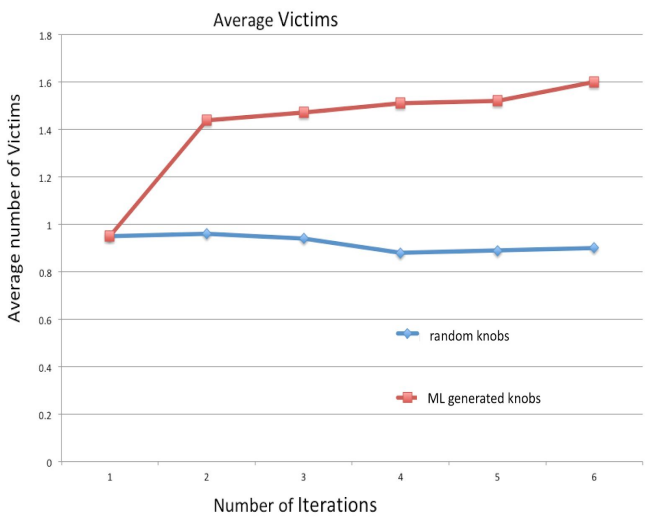

Fig.14   Victim buffer occupancy for random instruction generation versus ML generated  settings

The results from ML based optimization (orange curve) are compared with random testing (blue curve) in Figure 14 above. For each subsequent iteration through ML based optimization the average victim occupancy increases.

Figure 15 shows a histogram of the number of occurrences of each victim occupancy range for the original set of random runs and Figure 16 shows the third iteration of ML optimization. It is clear from these histograms than ML optimization has resulted in a shift to higher VB occupancy with the average VB occupancy increasing from 0.95 to 1.471 after just three iterations, a gain of almost 55%.

These results demonstrate that ML based optimization can scale to complex designs and help solve real-world DV problems.

CONCLUSIONS

We have demonstrated that using Machine Learning (Supervised and Reinforcement Learning) can significantly reduce the time to find hard to reach conditions in verification of complex IC's by generating input stimulus that performs significantly better than constrained-random inputs.

The cost of verifying hardware and software is increasing significantly with each generation of devices. The number of chips being produced is increasing due to the exponential growth in connected devices (IOT) and Machine Learning Applications.
The number of mobile apps being produced and software content generally are growing exponentially.
It is more critical now than ever to automate software and hardware verification to test hard to hit conditions in order to make software and hardware work in a safe predictable manner.

Machine Learning based stimulus generation can do better than random and has significant implications on other areas, such as verification of Android and IOS applications, network verification, pharmaceutical drug discovery, new material discovery etc. For example: A popular Android testing framework [19] that generates Random UI events is widely used to test Android mobile applications. Using a Machine Learning based approach as we have described, to generate better than random UI events, can significantly improve the efficiency of the Android UI testing framework.

Machine Learning based stimulus generation can lead to a dramatic increase in efficiency for discovering hardware and software bugs. Any simulation environment that uses random or constrained random methods to generate stimulus, such as drug discovery, Network verification etc, can benefit significantly by using the Machine Learning techniques presented in this paper.

We believe that Machine Learning based verification is the most feasible path forward to address the growing complexity of verifying software and hardware systems.

ACKNOWLEDGMENTS
We would like to thank Professor Pieter Abbeel of University of California Berkeley[15][16][17] for his valuable advice and deep insights into Reinforcement Learning.

We would like to thank Professor Jacob Abraham of University of Texas Austin,  (Department of Electrical Engineering and Computer Science) who provided valuable guidance and advice towards making the experiments a success.

We would  also like to thank Dr. Woody Kober of INZONE.AI who provided a rigorous mathematical framework to validate our hypothesis, algorithms and experiments.